%% file: main.tex
\documentclass[9pt, conference]{IEEEtran}
\usepackage{cite}
\usepackage{amsmath,amssymb,amsfonts}
\usepackage{algorithmic}
\usepackage{graphicx}
\usepackage{textcomp}
\usepackage[table]{xcolor}
\usepackage{threeparttable}
\usepackage{booktabs}
\usepackage{url}
\usepackage{comment}
\usepackage{adjustbox}
\newcolumntype{M}[1]{>{\centering\arraybackslash}m{#1}}
\usepackage[ruled,vlined]{algorithm2e}
\usepackage{caption}
\usepackage{subcaption}
\DeclareCaptionFormat{customcaptionsizing}{\fontsize{8}{8}\selectfont#1#2#3}
\def\BibTeX{{\rm B\kern-.05em{\sc i\kern-.025em b}\kern-.08em
    T\kern-.1667em\lower.7ex\hbox{E}\kern-.125emX}}

\makeatletter
\def\ps@IEEEtitlepagestyle{
  \def\@oddfoot{\mycopyrightnotice}
  \def\@evenfoot{}
}
\def\mycopyrightnotice{
  {\footnotesize
  \begin{minipage}{\textwidth}
  \centering
  \copyright~2021 IEEE.  Personal use of this material is permitted.  Permission from IEEE must be obtained for all other uses, in any current or future media, including reprinting/republishing this material for advertising or promotional purposes, creating new collective works, for resale or redistribution to servers or lists, or reuse of any copyrighted component of this work in other works. DOI: 10.1109/DAC18074.2021.9586094
  \end{minipage}
  }
  \gdef\mycopyrightnotice{}
}
\makeatother

\begin{document}

\title{PrefixRL: Optimization of Parallel Prefix Circuits using Deep Reinforcement Learning
}

\author{\IEEEauthorblockN{Rajarshi Roy, Jonathan Raiman, Neel Kant, Ilyas Elkin, Robert Kirby, \\
Michael Siu, Stuart Oberman, Saad Godil, Bryan Catanzaro}
\IEEEauthorblockA{\textit{NVIDIA}, Santa Clara, CA, USA\\
\{rajarshir, jraiman, nkant, ielkin, rkirby, msiu, soberman, sgodil, bcatanzaro\}@nvidia.com}}

\maketitle

\input{0_abstract}
\input{1_intro}
\input{2_background}
\input{3_implementation}
\input{4_results}
\input{5_conclusion}

{
\small
\bibliographystyle{IEEEtran}
\bibliography{references}
}

\end{document}

%% file: 0_abstract.tex
\begin{abstract}
In this work, we present a reinforcement learning (RL) based approach to designing parallel prefix circuits such as adders or priority encoders that are fundamental to high-performance digital design. Unlike prior methods, our approach designs solutions tabula rasa purely through learning with synthesis in the loop. We design a grid-based state-action representation and an RL environment for constructing legal prefix circuits. Deep Convolutional RL agents trained on this environment produce prefix adder circuits that Pareto-dominate existing baselines with up to 16.0\% and 30.2\% lower area for the same delay in the 32b and 64b settings respectively. We observe that agents trained with open-source synthesis tools and cell library can design adder circuits that achieve lower area and delay than commercial tool adders in an industrial cell library.
\end{abstract}


%% file: 1_intro.tex
\section{Introduction}
\label{sec:introduction}
Several fundamental digital design building blocks such as adders, priority encoders, inc(dec)rementers and gray-to-binary code converters can be reduced to prefix-sum computations and implemented as (parallel) prefix circuits \cite{lin2011introduction}. Thus, the optimization of prefix circuits for area, delay and power is an important and well studied problem in digital hardware design.

The optimization of prefix circuits is challenging as their large design space grows exponentially with input length and is intractable to enumerate. As a result, exhaustive search approaches do not scale beyond small input lengths \cite{verma2006towards}. Several regular prefix circuit structures \cite{sklansky1960conditional, kogge1973a, brent1982a} have been proposed that trade off logic level, maximum fanout and wiring tracks. Another set of algorithms \cite{matsunaga2007area, liu2003an, fishburn1990a, zimmermann1996non} optimize prefix circuit size and level properties. However, \cite{ma2019cross} observes that prefix circuit level and maximum fanout properties do not map to circuit area, power and delay due to physical design complexities such as capacitive loading and congestion.

\begin{figure}[htbp]
    \setlength{\abovecaptionskip}{3pt}
    \setlength{\belowcaptionskip}{-10pt}
\centerline{\includegraphics[width=\linewidth]{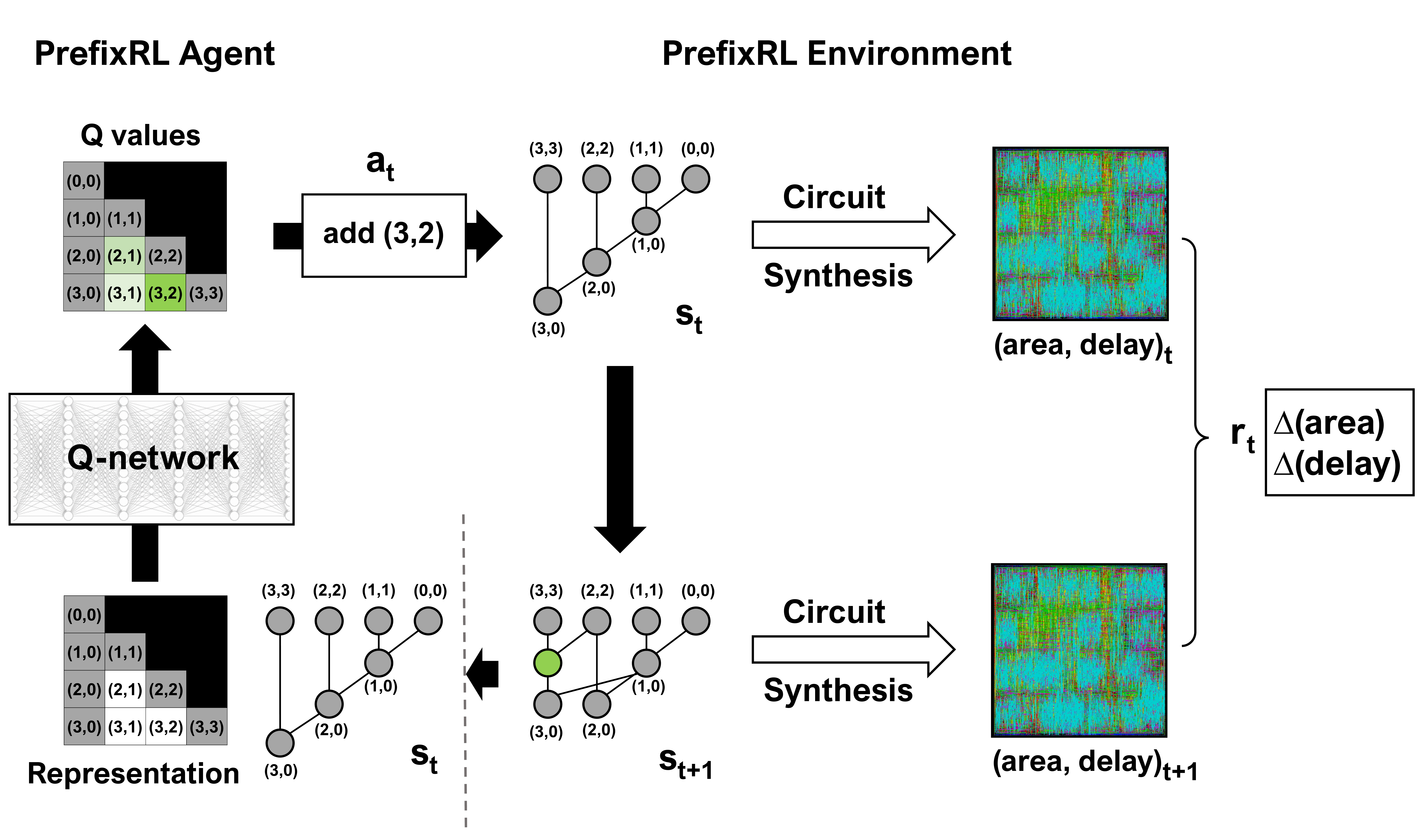}}
\caption{PrefixRL flow. $s_t$ shown is the ripple-carry prefix graph, a possible starting state $s_0$. The Q-network takes action (3,2) modifying the circuit and receiving a reward computed by the area/delay difference in the circuits corresponding to $s_t$ and $s_{t+1}$. Refer to Section \ref{sec:trainsystem} for circuit synthesis details. For clarity we show a 4b circuit but our primary results are obtained in the 32b and 64b setting.}
\label{fig:flow}
\end{figure}

Meanwhile, there are a growing number of success stories in other domains using deep reinforcement learning (RL) \cite{sutton1998rl} to produce sophisticated solutions to sequential decision problems. RL agents have outperformed humans in complex games \cite{silver2018a} and produced novel solutions to search and design problems \cite{mirhoseini2020chip}. PrefixRL continues this trend in the circuit design domain. The key contributions of this work are:
\begin{itemize}
\item An RL framework that trains an agent, tabula rasa, to explore the unrestricted prefix circuit space with synthesis in the loop (Fig. \ref{fig:flow}) while optimizing for area and delay.
\item We demonstrate the capabilities of PrefixRL with the task of optimizing 32b and 64b prefix adder circuits. PrefixRL explores the massive $\mathcal{O}(2^{N^2})$ design space and generates state-of-the-art frontiers of designs that Pareto-dominate all the designs found by prior work using simulated annealing \cite{moto2018prefix}, exhaustive search with pruning \cite{roy2014towards}\cite{ma2019cross} and regular structures \cite{sklansky1960conditional, kogge1973a, brent1982a} when synthesized in an industrial process technology, achieving a maximum area savings of 16.0\% and 30.2\% for equivalent delay targets in the 32b and 64b settings.
\item We observe that agents trained with open-source synthesis tools (OpenPhySyn \cite{openphysyn}) and cell library (Nangate45 \cite{nangate}) design adder circuits that can achieve lower area and delay than commercial tool adders in an industrial cell library (8nm).
\item We illustrate the challenges of training an RL agent with continuous synthesis feedback and the solutions required to handle workloads in similar use cases. With our state-of-the-art results, these solutions provide a compelling blueprint for future RL approaches to design automation.
\end{itemize}

\section{Related Work}
\label{sec:relatedwork}

Several approaches attempt to optimize the prefix circuits directly for area, power and delay. \cite{roy2014towards} utilizes a combination of heuristic rules to prune the intractable design space of 8, 16, 32 and 64 bit adders to a small subset that can be exhaustively searched. \cite{ma2019cross} extends this technique by proposing an alternative set of pruning heuristics that result in a larger set of pruned adders which are then searched using a machine learning model that is trained to predict physical metrics. Our approach is fundamentally different as it does not rely on any hand-crafted heuristics. Our agents learn tabula rasa to explore the full prefix adder space with synthesis in the loop.

Another work \cite{moto2018prefix} uses simulated annealing (SA) \cite{kirkpatrick1983optimization} to randomly modify and legalize 32b prefix adder graphs towards optimal structures in an unrestricted design space evaluated using an analytical model. They obtain a frontier of designs that optimize these competing metrics more effectively than the regular structures \cite{sklansky1960conditional, kogge1973a, brent1982a} as well as those obtained by the previously mentioned exhaustive search on a heuristically pruned search space \cite{roy2014towards}. However, all of these results are limited to analytical evaluation metrics. Our prefix RL approach instead optimizes physical synthesis metrics directly with synthesis in the loop. We investigate applying our approach with the analytical evaluation metrics and demonstrate that our prefix circuits Pareto dominate the SA based baseline circuits (Section \ref{sec:importancesynthesis}). However, we observe that these circuits that were found using the analytical evaluation metric significantly degrade in quality once they go through physical synthesis. Note that since physical synthesis is significantly more computationally intensive than the analytical evaluation, it would not be feasible to scale a fundamentally sequential algorithm such as SA to use physical synthesis in the loop.

%% file: 2_background.tex
\section{Background}
\label{sec:background}
\subsection{Prefix Graphs}
The generalized prefix-sum computation is to compute $y_i = x_i\circ x_{i-1}\circ\cdot\cdot\cdot\circ x_0$ for $0\leq i \leq N-1$, given $N$ inputs $x_0,x_1,...,x_{N-1}$ and a binary associative operator $\circ$ \cite{blelloch1990prefix}.

An N-input prefix-sum computation can be performed in several ways due to the associativity of the operator. For example, two of the ways the 4-input prefix sum can be computed are:
\begin{multline*}
y_0=x_0, y_1=x_1\circ y_0, y_2=x_2\circ y_1, y_3=x_3\circ y_2\\
y_0=x_0, y_1=x_1\circ y_0, y_2=x_2\circ y_1, z_{3:2}=x_3\circ x_2, y_3=z_{3:2}\circ y_1
\end{multline*}

Introducing the additional term $z_{3:2}$, breaks the dependency of $y_3$ on $y_2$ and allows it to be computed in parallel with $y_2$, thus the term parallel prefix \cite{ladner1980parallel}. In general, we denote $z_{i:j}$ to represent $x_i\circ x_{i-1}\circ\cdot\cdot\cdot\circ x_j$. Then the outputs $y_i$ can be rewritten as $z_{i:0}$ and inputs $x_i$ can be rewritten as $z_{i:i}$. Note that $y_0$ and $x_0$ both correspond to $z_{0:0}$

Parallel prefix computations can be represented as a directed acyclic prefix graph where every computation unit $z_{i:j}$ is a graph node that performs a single operation on a pair of inputs: $z_{i:j} = z_{i:k}\circ z_{k-1:j}$ where $i \geq k > j$.
We use the notation from \cite{roy2014towards} where the most and least significant bits ($MSB$, $LSB$) of computation node $z_{i:j}$ is $(i, j)$. Using this notation we will term the node $(i, k)$ as the \textit{upper parent} of $(i, j)$ and the node $(k-1, j)$ as its \textit{lower parent}. The prefix graphs corresponding to the 4-input prefix sum computations above are shown in Fig.~\ref{fig:flow} as $s_t$ and $s_{t+1}$ . In both graphs, the upper and lower parents of node $(2,0)$ are $(2,2)$ and $(1,0).$

Every legal $N$-input prefix graph must have input nodes $(i, i)$, output nodes $(i, 0)$ for $1\leq i \leq N-1$, and the input/output node $(0,0)$. Furthermore, every non-input node must have exactly one upper parent ($up$) and one lower parent ($lp$) such that:
\begin{align} \label{prefixrules}
LSB(node) &= LSB(lp(node)) \nonumber\\
LSB(lp(node)) &\leq MSB(lp(node)) \nonumber\\
MSB(lp(node)) &= LSB(up(node))-1 \nonumber\\
LSB(up(node)) &\leq MSB(up(node)) \nonumber\\
MSB(up(node)) &= MSB(node)
\end{align}

\subsection{Deep Reinforcement Learning}

Reinforcement learning (RL) \cite{sutton1998rl} is a class of algorithms applicable to sequential decision making tasks. RL makes use of the Markov Decision Process (MDP) formalism wherein an \textit{agent} attempts to optimize a function in its \textit{environment}. An MDP can be completely described by a \textit{state space} $\mathcal{S}$ (with states $s \in \mathcal{S}$), an \textit{action space} $\mathcal{A}$ (with actions $a \in \mathcal{A}$), a \textit{transition function} $\mathcal{T}: \mathcal{S} \times \mathcal{A} \rightarrow \mathcal{S}$ and a \textit{reward function} $\mathcal{R}: \mathcal{S} \times \mathcal{A} \rightarrow \mathbb{R}$. 
In an MDP, an episode evolves over discrete timesteps $t=0,1,2,..$ where the agent observes $s_t$ and responds with action $a_t$ using a \textit{policy} $\pi(a_t|s_t)$. The environment provides to the agent the next state $s_{t+1} = \mathcal{T}(s_t, a_t)$ and the reward $r_t = \mathcal{R}(s_t, a_t)$. The agent is tasked with maximizing the \textit{return} (cumulative future rewards) by learning an optimal policy $\pi^*$.

The $Q$ value of a state-action pair $(s_t, a_t)$ under a policy $\pi$ is defined to be the expected return if action $a_t$ is taken at state $s_t$ and future actions are taken using the policy $\pi$.
\begin{equation}
Q^\pi(s_t,a_t) = \mathbb{E}[r_t+\gamma r_{t+1}+\gamma^2 r_{t+2}+\cdot\cdot\cdot], \gamma\in[0,1]
\end{equation}
The discount factor $\gamma\in [0,1]$ balances short-term versus long-term rewards. The $Q$-learning algorithm \cite{watkins1992q} starts the agent with a random policy and uses the experience gathered during its interaction with the environment $(s_t,a_t,r_t,s_{t+1})$ to iterate towards an optimal policy by updating $Q$ with a learning rate $\alpha \in \left[0, 1\right]$:
\begin{equation}
Q(s_t,a_t) \leftarrow (1 - \alpha) * Q(s_t,a_t) + \alpha * (r_t+\gamma \max\limits_{a^{\prime}} Q(s_{t+1},a^{\prime}))
\end{equation}
The policy for a $Q$-learning agent is simply $\pi(\cdot|s_t) = \underset{a}{\operatorname{argmax}} \: Q(s_t,a)$. We use the $\epsilon$-greedy policy, where random actions are chosen with a probability $\epsilon$ to increase exploration in the state space. $\epsilon$ is annealed to zero during the course of training and is always zero when doing evaluation.

The DQN (deep $Q$ network) algorithm \cite{mnih2015human} uses a deep neural network as a $Q$ value function approximator to achieve human-level performance on several Atari games. DQN stabilizes training using a second \textit{target network} to estimate the $Q$ values of $(s_{t+1}, a^\prime)$ that is updated less frequently and sampling an \textit{experience replay} buffer. The Double-DQN algorithm \cite{hasselt2016deep} further improves training by reducing harmful overestimations in DQN.

%% file: 3_implementation.tex
\section{PrefixRL Implementation}
\label{sec:methodology}
We frame the optimization of prefix circuits as a RL task by creating an MDP for their construction. We pick prefix adders due to their importance in arithmetic datapaths and focus on minimizing circuit area and delay. We then train multiple PrefixRL agents to design an area-delay minimized Pareto frontier of adders.

\subsection{Reinforcement Learning Environment}
\label{sec:rlenv}
The PrefixRL state space $\mathcal{S}$ consists of all legal $N$-input prefix graphs. $N$-input graphs can be represented in a $N \times N$ grid with rows representing $MSB$ and columns representing $LSB$ (Fig.~\ref{fig:flow}). Note that the input nodes ($MSB=LSB$) will lie on the diagonal, output nodes will lie on the first column ($LSB=0$) and locations above the diagonal ($LSB>MSB$) cannot contain a node. The remaining $(N-1)(N-2)/2$ locations where non-input/output nodes may or may not exist define the $\mathcal{O}(2^{(N-1)(N-2)/2}) = \mathcal{O}(2^{N^2})$ state space of $N$-input prefix graphs. For example, 32-input graphs will have $\vert \mathcal{S} \vert = \mathcal{O}(2^{465})$ with a lower exact value because not all combinations of nodes in those locations will meet the legality constraints in (\ref{prefixrules}).

The action space $\mathcal{A}$ for an $N$-input prefix graph consists of two actions (add or delete) for any non-input/output  node i.e. where $LSB\in [1,N-2]$ and $MSB\in [LSB+1, N-1]$. Hence, $\vert \mathcal{A} \vert = (N-1)(N-2)/2$. The environment evolution through $\mathcal{T}$ always maintains a legal prefix graph by:
\begin{enumerate}
\item Applying a legalization procedure after an action that may add or delete additional nodes to maintain legality.
\item Forbidding redundant actions that gets undone by the legalization procedure.
\end{enumerate}
During legalization, the upper parent of a node, $up(node)$, is the existing node with same $MSB$ and the next highest $LSB$. The lower parent of a node is computed using the node and its upper parent (\ref{prefixrules}): \[(MSB_{lp(node)}, LSB_{lp(node)}) = (LSB_{up(node)}-1, LSB_{node})\]
An illegal condition happens only when the lower parent $lp(node)$ of a node does not exist, so the legalization procedure adds any missing lower parent nodes.

\input{algo_ppg}

The action of adding a node that already exists (in $nodelist$) is redundant and is forbidden. Deleting is limited to nodes in $minlist$ (nodes that are not lower parents of other nodes) to prevent legalization from adding back deleted nodes (Algorithm~\ref{algo:pgg}).

\subsection{Scalarized Double Deep Q Learning}
\label{sec:scalarddqn}
PrefixRL uses a scalarized version of the Double-DQN algorithm \cite{hasselt2016deep}. Every episode starts the environment with the initial state $s_0$ randomly chosen to be either a ripple-carry or a Sklansky \cite{sklansky1960conditional} graph. These are prefix graphs with the minimum node and level count respectively. Every action $a_t$ from the agent modifies the legal prefix graph $s_t$ to another legal prefix graph $s_{t+1}$ and returns a reward vector $\mathbf{r_t}$ that indicates the decrease in the normalized circuit area and delay when its prefix graph is modified from $s_t$ to $s_{t+1}$ (Fig.~\ref{fig:flow}). Details of how we measure area and delay are given in Section \ref{sec:trainsystem}.
\begin{align} \label{rewardcalc}
\mathbf{r_t} =[area(s_{t})-area(s_{t+1}), delay(s_{t})-delay(s_{t+1})] \nonumber
\end{align}

With competing objectives such as area and delay, the same improvement in the scalarized objective can occur from either an improvement in area or delay, but the resulting prefix graph structures would be very different. Thus, it is difficult for a Q-learning agent to infer how its actions affect prefix graphs if it can only observe a scalarized objective. The \textit{scalarized deep Q-learning} algorithm \cite{mossalam2016multi} updates the $Q$-learning algorithm for such multi-objective settings by receiving rewards and learning the Q value for different objectives separately but choosing actions after scalarizing the $Q$ values with a weight vector $\mathbf{w}$.

The PrefixRL agent estimates the Q function vector $\mathbf{Q}(s,a) = [Q_{area}(s,a), Q_{delay}(s,a)]$ for a pair of prefix graph state and modification action using a deep neural network. The double-DQN training procedure \cite{hasselt2016deep} is used with the target, loss and action selection extended for scalarization:
\begin{equation} \label{qtarget}
\mathbf{y_t} = \mathbf{r_t}+\gamma\mathbf{Q}(s_{t+1}, \underset{a}{\operatorname{argmax}}[\mathbf{w}^\top \mathbf{Q}(s_{t+1},a; \theta _t)];\theta'_t)
\end{equation}
\begin{equation} \label{qloss}
Loss(\mathbf{Q}(s_t,a_t;\theta_t), \mathbf{y_t})
\end{equation}
\begin{equation} \label{actionselect}
a_t = \underset{a}{\operatorname{argmax}}[\mathbf{w}^\top\mathbf{Q}(s_t,a; \theta_t)]
\end{equation}
Where $\theta_t$ and $\theta'_t$ are the parameters of the local and target networks in the double-DQN algorithm.

A Pareto frontier of designs can then be obtained by solving multiple single-objective optimization problems, each with the scalarized objective $\mathbf{w}^\top \mathbf{m} = w_{area}\cdot area \: + \: w_{delay}\cdot delay$ using different scalarization weights $\mathbf{w}$.  We conceptually encode a tradeoff of objectives by normalizing $\mathbf{w}$ such that its elements are nonnegative and sum to 1. However, we must also scale the raw values of $area$ and $delay$ since their units are incomparable. Our procedure for this is to multiply those values by scaling constants $c_{area}, c_{delay}$ such that the Pareto frontier for different $\mathbf{w}$ evenly covers the breadth of baseline prefix graph designs synthesized with multiple delay targets. In our experiments we use $c_{area} = 0.001$ and $c_{delay} = 10$.

\subsection{Q Network Architecture}
\label{sec:dqnarch}
The deep neural network $Q$ value approximator takes the state $s_t$ as the input and predicts  $\forall a \in \mathcal{A}: [Q_{area}(s_t, a),Q_{delay}(s_t, a)]$.
Based on the $N \times N$ grid based representation of prefix graphs described in Section~\ref{sec:rlenv}, the input to the neural network is a $N\times N \times 4$ tensor where the 4 channels encode node features as:
\begin{enumerate}
    \item 1 if node $(MSB, LSB)$ in $nodelist$, 0 otherwise
    \item 1 if node $(MSB, LSB)$ in $minlist$, 0 otherwise
    \item level of node $(MSB, LSB)$ in $nodelist$, 0 otherwise
    \item fanout of node $(MSB, LSB)$ in $nodelist$, 0 otherwise

\end{enumerate}
where the fanout of a node refers to the number of children it has and the level of a node refers to its topological depth from input nodes in the prefix graph. Features are normalized to $\left[0,1\right]$.

\begin{figure}[htbp]
\centerline{\includegraphics[width=\linewidth]{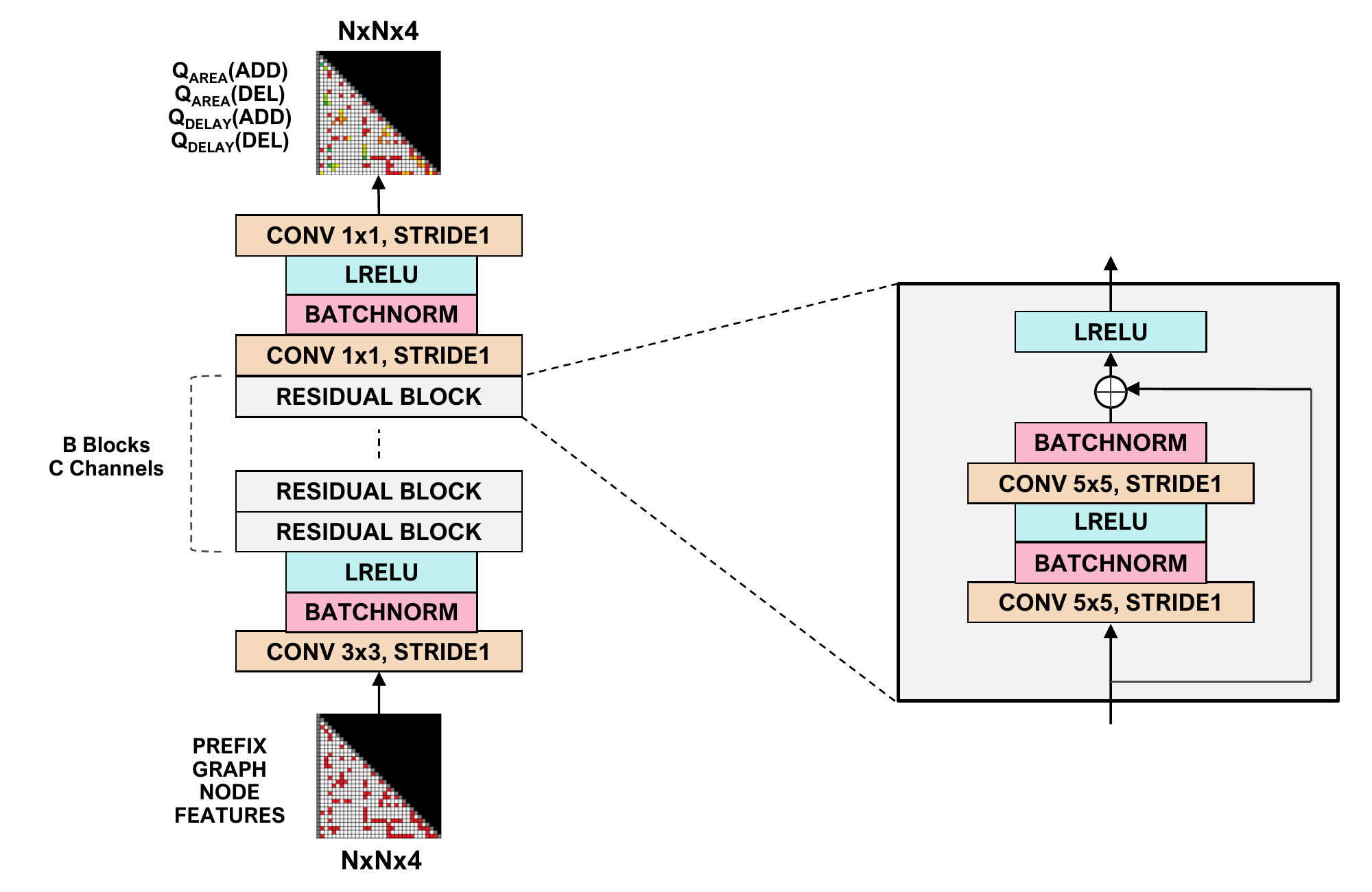}}
\caption{$Q$-network architecture for $N$-input prefix graphs. For both 32b and 64b, our architecture uses $B = 32$ and $C = 256$.}
\label{fig:netarch}
\end{figure}

The PrefixRL agent uses a convolutional architecture \cite{krizhevsky2012imagenet} in a residual network \cite{he2016deep} configuration similar to \cite{silver2018a}. The neural network consists of a ``body" followed by a $Q$-value ``head", with the details of the layer types and connections given in Fig. \ref{fig:netarch}. The output is a $N\times N \times 4$ matrix where the 4 channels represent the $Q_{area}$ and $Q_{delay}$ values for adding and for deleting the node $(MSB, LSB)$. We use $nodelist$ and $minlist$ to set the Q values of illegal actions to $-\infty$ so that they are never chosen. In our experiments, we use a discount factor of $\gamma = 0.75$, an experience buffer with up to $4 \times 10^5$ elements, synchronize our target $Q$-network every 60 gradient steps and train using the Adam optimizer with learning rate $4 \times 10^{-5}$. 

\subsection{Training System Description} 
\label{sec:trainsystem}

\begin{figure}[htbp]
\centerline{\includegraphics[width=\linewidth]{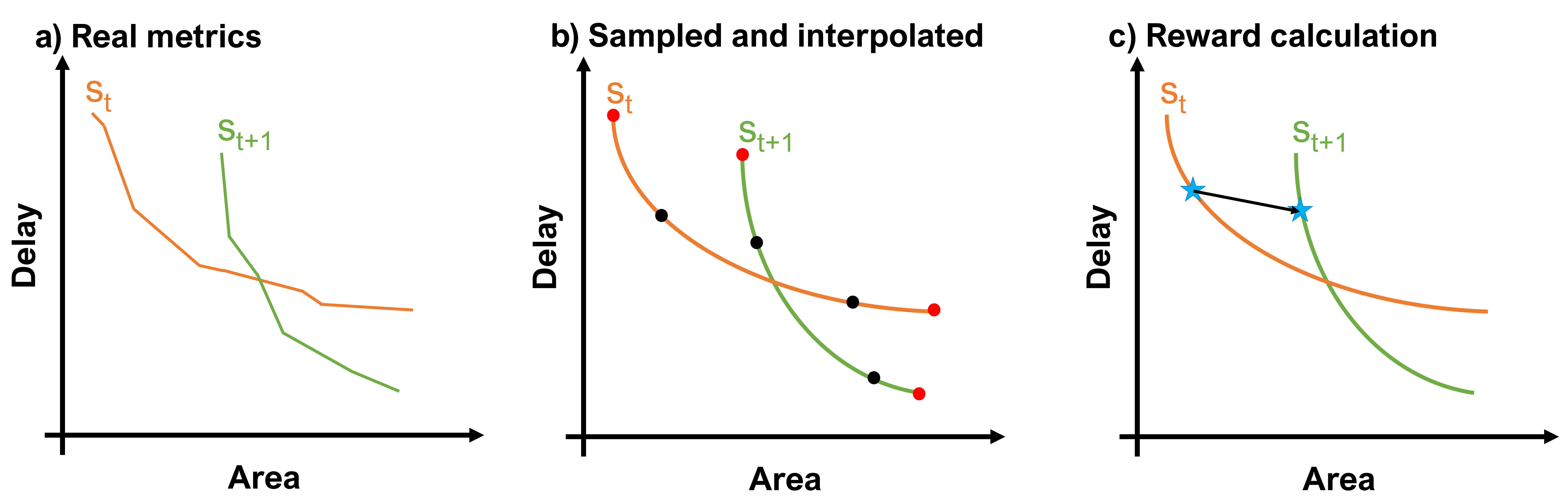}}
\caption{Process for calculating reward in PrefixRL. (a) Each prefix state corresponds to a curve of synthesized circuit metrics based on physical synthesis timing constraints. (b) We only sample 4 such points and interpolate a curve for the metrics. (c) The reward is then calculated as the vector difference between the \textbf{w}-optimal points on the curves from $s_{t}$ and $s_{t+1}$.}
\label{fig:rewardcalc}
\end{figure}

Building the deep learning training system for PrefixRL is a nontrivial task due to the time and resource-intensiveness of the reward calculation $\mathcal{R}(s, a)$. For each prefix graph state we perform circuit synthesis by generating a gate level netlist (Section \ref{sec:adderopt}) and applying timing-driven synthesis optimizations at 4 delay targets. We use the OpenPhySyn physical synthesis tool \cite{openphysyn} for optimizations such as gate sizing, gate cloning, buffer insertion and pin swapping to prefix circuits. Section \ref{sec:importancesynthesis} highlights the importance of these optimizations. After synthesis optimization, we interpolate an area-delay tradeoff curve using PChip Interpolation (Fig.~\ref{fig:rewardcalc}). Processing a single state with OpenPhySyn takes up to 36 seconds on average for the 64b case (see Table \ref{fig-scaling}) and RL training runs usually span on the order of at least $10^5$ environment steps so it is necessary to parallelize synthesis as much as possible.

We implement an asynchronous distributed system to efficiently perform synthesis in the training loop. Our main training process launches additional processes on separate nodes with large CPU resources to run the OpenPhySyn workload and returns the final area and delay information. This asynchronous workflow is made efficient through pipeline parallelism which hides the latency of individual calls and amortizes the overall delay to be asymptotically minimal. Furthermore, we cache synthesized state designs to reduce redundant calculations and find that as the exploration parameter $\epsilon$ diminishes, the cache hit percentage becomes 50\% in the 32b case and 10\% in the 64b case.

Aside from parallelizing synthesis in the loop, we also built a distributed framework for efficient RL. The key observation is that DQN is an \textit{off-policy} RL algorithm, meaning that training can be done on experience gathered by any policy, not simply the current version. This effectively decouples and allows for parallelizing the process of generating new experience and training on prior experience. For this to be viable, we must ensure an appropriate amount of new experience is generated with each new version of the parameters. We found that having 192 synthesis worker processes generating experience was empirically sufficient to satisfy this condition.

%% file: algo_ppg.tex
\begin{algorithm}[htbp]
\scriptsize

\SetAlgoLined
\SetKwFunction{FInit}{Initialize}
\SetKwFunction{FAdd}{Add}
\SetKwFunction{FDel}{Delete}
\SetKwFunction{FLeg}{Legalize}

\SetKwProg{Fn}{Function}{:}{}
\Fn{\FInit}{
 $nodelist \leftarrow \emptyset, minlist \leftarrow \emptyset$\;
 \For(\tcp*[f]{add in/out nodes}){$m\leftarrow 0$ \KwTo $(N-1)$}{
  add $(m,m)$, $(m,0)$ to $nodelist$
 }
}

\SetKwProg{Fn}{Function}{:}{}
\Fn{\FAdd{msb,lsb}}{
 add $(msb,lsb)$ to $minlist$\;
 \tcp{remove new node's and child's lps from minlist}
 \For{$l\leftarrow (msb-1)$ \KwTo $0$}{
  \If{$(msb,l)$ is in $minlist$}{
   delete $lp(msb,l)$ from $minlist$
  }
 }
 \FLeg{}
}

\SetKwProg{Fn}{Function}{:}{}
\Fn{\FDel{msb,lsb}}{
 delete $(msb,lsb)$ from $minlist$\;
 \FLeg{}
}
\SetKwProg{Fn}{Function}{:}{}
\Fn{\FLeg}{
 $nodelist \longleftarrow minlist$\;
 \For(\tcp*[f]{add in/out nodes}){$m\leftarrow 0$ \KwTo $(N-1)$}{
  add $(m,m)$, $(m,0)$ to $nodelist$
 }
 \For(\tcp*[f]{add missing lps}){$m\leftarrow (N-1)$ \KwTo $0$}{
  \For{$l\leftarrow (m-1)$ \KwTo $0$}{
   \If{$(m,l)$ is in $nodelist$}{
    add $lp(m,l)$ to $nodelist$
   }
  }
 }
}

\caption{\small PrefixRL N-input prefix graph actions}
\label{algo:pgg}
\end{algorithm}

%% file: 4_results.tex
\section{Results and Discussion}
\label{sec:results}

\subsection{Prefix Adder Optimization with OpenPhySyn}
\label{sec:adderopt}
With the objective of minimizing the synthesized area and delay of prefix adder circuits, multiple PrefixRL agents were trained with 15 area-delay scalarization weights $\mathbf{w}$ in the range [0.10, 0.99]. 
The prefix adders were generated from prefix graphs using alternating NAND/NOR, OAI/AOI, XNOR, NOR and INV gates in the Nangate45 cell library\cite{nangate} based on the implementation described in \cite{zimmermann1998binary}. The training approach described in Section \ref{sec:trainsystem} was used with the timing-driven synthesis optimizations in the OpenPhySyn\cite{openphysyn} synthesis tool. Since OpenPhySyn and Nangate45 are in the open-source domain, our training procedure and results are reproducible without requiring commercial tools.

We note that circuit power is an important metric that should ideally be jointly optimized with area and delay. However, due to the computational requirements of power simulation, we did not integrate this as a third objective. We leave the integration of a power objective to the optimization as future work. Furthermore, PrefixRL agents learn policies to design prefix circuits under uniform timing constraints since the circuit synthesis environment specify uniform arrival and departure times for inputs and outputs. An interesting future direction would be to train generalizable agents that adapt to various nonuniform timing constraints.

After training, the various PrefixRL agents learn to design adders specializing at various area-delay tradeoff points after considering synthesis optimizations. We synthesize the various adders generated by PrefixRL and baseline approaches at 40 delay targets with OpenPhySyn and Nangate45. Since each delay target potentially generates a different circuit for the same design, we bin all adder circuits for an approach and present the area-delay Pareto front.

\input{figtex/fig_openphysynresults}

The baselines for 32b prefix adders are regular Sklansky\cite{sklansky1960conditional}, Kogge-Stone\cite{kogge1973a}, Brent-Kung\cite{brent1982a} adders, and adders from simulated annealing (SA)\cite{moto2018prefix} and pruned search (PS)\cite{roy2014towards} approaches. (Fig.~\ref{fig:openphysynresults32}) shows that Prefix-RL agents learn to design 32b adders that Pareto dominate all these approaches. Throughout much of the delay targets ($\ge$ 30 ns), the percent improvement in area is consistent but only 2 to 8 percentage points. However, the gains become much more significant at lower targets, reaching a maximum area saving of \textbf{16.0\%} at delay target 0.293 ns.

The baselines for 64b prefix adders are the regular adders, and 1100 adders from the machine learning driven cross layer optimization approach (CL)\cite{ma2019cross}. (Fig.~\ref{fig:openphysynresults64}) shows that Prefix-RL agents produce 64b adders that Pareto dominate this work as well. Samples of the learnt adders are visible in Fig. \ref{fig:64solutions}. Compared to the 32b setting, we observe large gains over the baselines in the knee of the curve. This illustrates the power of an RL approach as it can successfully scale to larger problem sizes in ways that other unrestricted search space algorithms like SA cannot. In this region, PrefixRL consistently yields area savings of 12 to 20 percentage points. At lower targets, PrefixRL does even better, achieving a maximum improvement of \textbf{30.2\%} at delay target 0.347 ns.

\subsection{Generalization to Industrial Settings}
To study the generalization of prefix circuit optimizations across cell libraries, we picked 7 Pareto-optimal PrefixRL adders and synthesized them at an 8nm industrial cell library with a commercial physical synthesis tool from a leading EDA vendor. In this setting, the adder is instantiated with inputs arriving from, and outputs feeding to flip-flops to ensure input uniform arrival and output departure times. We also allow the INV, XOR, XNOR cells in the adder netlist to undergo logic synthesis and technology mapping. We limit our comparison to regular adders and the library of adders instantiated by the tool (Commercial) and synthesize all the adders at the same 12 delay targets. We measure the cell area of just the adder in our results. (Fig.~\ref{fig:commercialresults32}) shows that Prefix-RL adders circuits Pareto-dominate the Kogge-Stone and Brent-Kung adders, while achieving lower area and delay than the Commercial and Sklansky adders in all the instances except the lowest delay target. This indicates that prefix circuit optimizations can generalize across cell libraries to a certain extent. We note that while training PrefixRL directly with the industrial cell library and synthesis tool may improve results further, there are possible design scenarios where PrefixRL adders may already yield better results than Commercial adders.

\input{figtex/fig_commercialresults}

\subsection{Scaling}
As discussed in Section \ref{sec:trainsystem}, PrefixRL requires considerable engineering optimization in order to be a tractable solution. We can quantify this improvement by considering that 64b results were obtained after $5.0 \times 10^5$ environment steps which took approximatedly 5 days of training on our parallelized infrastructure. To obtain this with a single-threaded version of PrefixRL would take over 8 times longer or about 44 days of training. 

\input{figtex/fig_scaling}

Even with these large efficiency gains, we still had to make concessions when scaling PrefixRL to the 64b setting. The larger state representation prevents us from further expanding our 64b model capacity, so we kept it equal to that of the 32b model, while we leverage data parallelism across multiple GPUs to fit training batches in GPU memory (Table \ref{fig-scaling}). Training also takes roughly twice as many environment steps as needed as our 32b models to produce the results in Fig. \ref{fig:openphysynresults32} and \ref{fig:commercialresults32}. Other common RL workloads, however, regularly reach the $10^6$ to $10^7$ environment step range, and in that context our solution is relatively data efficient while producing state-of-the-art results.

\subsection{Importance of Synthesis-In-The-Loop}
\label{sec:importancesynthesis}
While we primarily focus on the physical synthesis metrics since they correspond closely to real-world performance, adder prefix graph structures are also commonly evaluated with analytical metrics. We use the analytical evaluation to assess the advantage RL provides over existing approaches as well as to examine how well performance transfers between the analytical and physical synthesis settings. 

\input{figtex/fig_analyticalresults}

In Fig. \ref{fig:32anaresultsana}, we directly compare PrefixRL against a simulated annealing (SA) \cite{moto2018prefix} approach that also operates on the unrestricted design space of all prefix graphs. We trained multiple PrefixRL agents with various area-delay scalarization weights $\mathbf{w}$ and the same analytical models of node area ($1.0$) and delay ($1.0+0.5\cdot \mathrm{fanout}$) provided by \cite{moto2018prefix}. The trained Analytical-PrefixRL agents learn to produce 32b prefix graphs that Pareto-dominate (Fig.~\ref{fig:32anaresultsana}) all the published solutions from \cite{moto2018prefix} with \textbf{11.7\%} lower area at the lowest delay point, showing that RL itself can out-compete existing approaches in a setting that does not require expensive physical synthesis feedback. 

In order to compare the analytical metrics of prefix graphs against real circuit properties, we generated prefix adder circuits from the graphs and applied timing-driven synthesis optimizations at 40 target delays with OpenPhySyn and Nangate45. (Fig.~\ref{fig:32anaresultssynth}) shows that even though the Analytical-PrefixRL and SA designs Pareto-dominate the regular prefix graphs \cite{sklansky1960conditional, kogge1973a, brent1982a} and pruned search (PS) designs \cite{roy2014towards} at analytical metrics, they do not yield well to synthesis optimizations. The PS and Sklansky adders can achieve lower delay while maintaining lower area than the Analytical-PrefixRL and SA adders. These results highlight the importance of training directly with synthesized circuit area and delay objectives.

%% file: figtex/fig_openphysynresults.tex
\begin{figure}
    \setlength{\abovecaptionskip}{3pt}
    \setlength{\belowcaptionskip}{-10pt}
    \centering
    \begin{subfigure}[b]{\linewidth}
        \setlength{\abovecaptionskip}{0pt}
        \setlength{\belowcaptionskip}{0pt}
        \centering
        \includegraphics[width=\textwidth]{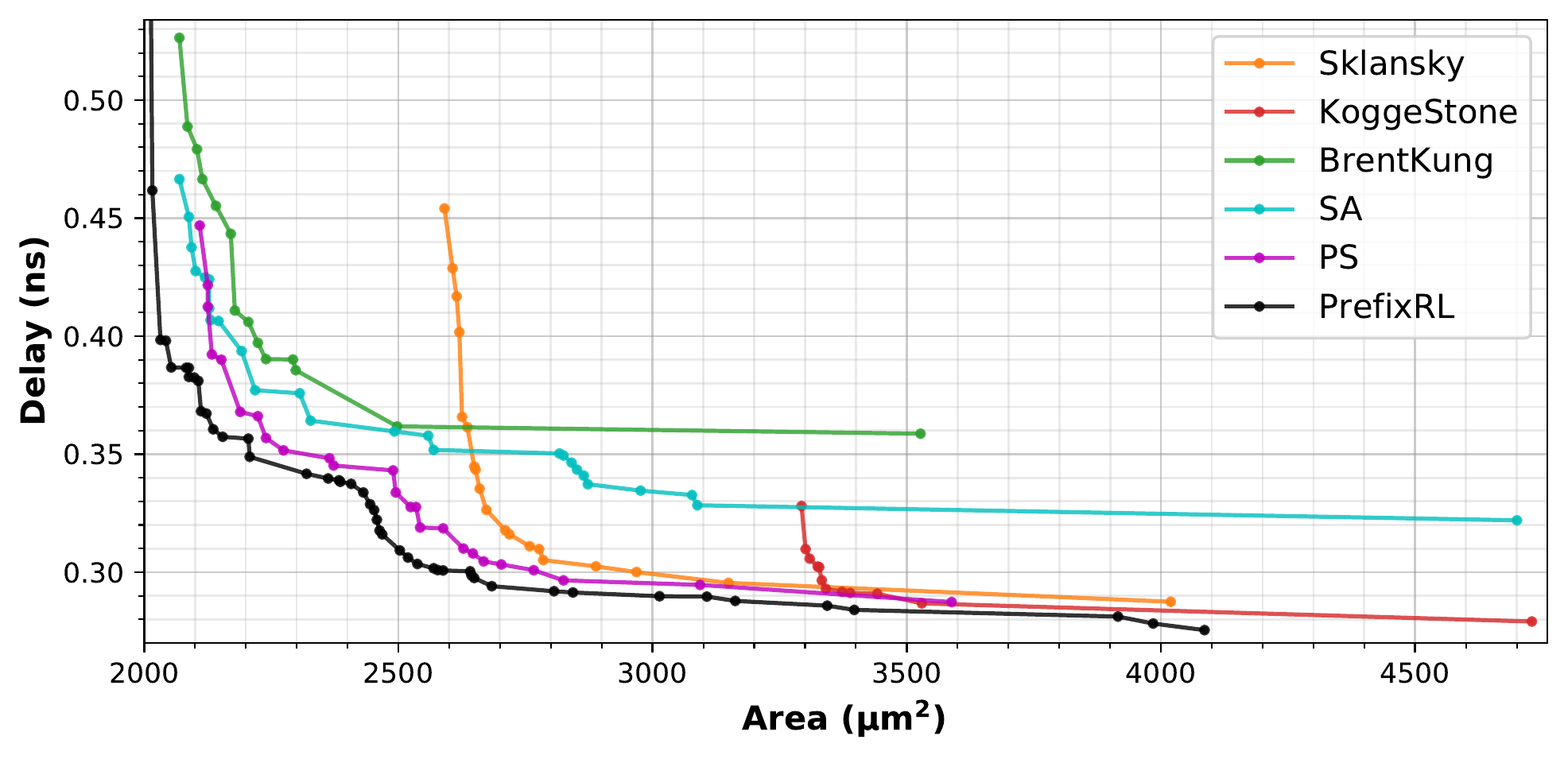}
        \caption{32b Adder Synthesis. OpenPhySyn, Nangate45.}
        \label{fig:openphysynresults32}
    \end{subfigure}%
    \hfill%
    \begin{subfigure}[b]{\linewidth}
        \setlength{\abovecaptionskip}{0pt}
        \setlength{\belowcaptionskip}{0pt}
        \centering
        \includegraphics[width=\textwidth]{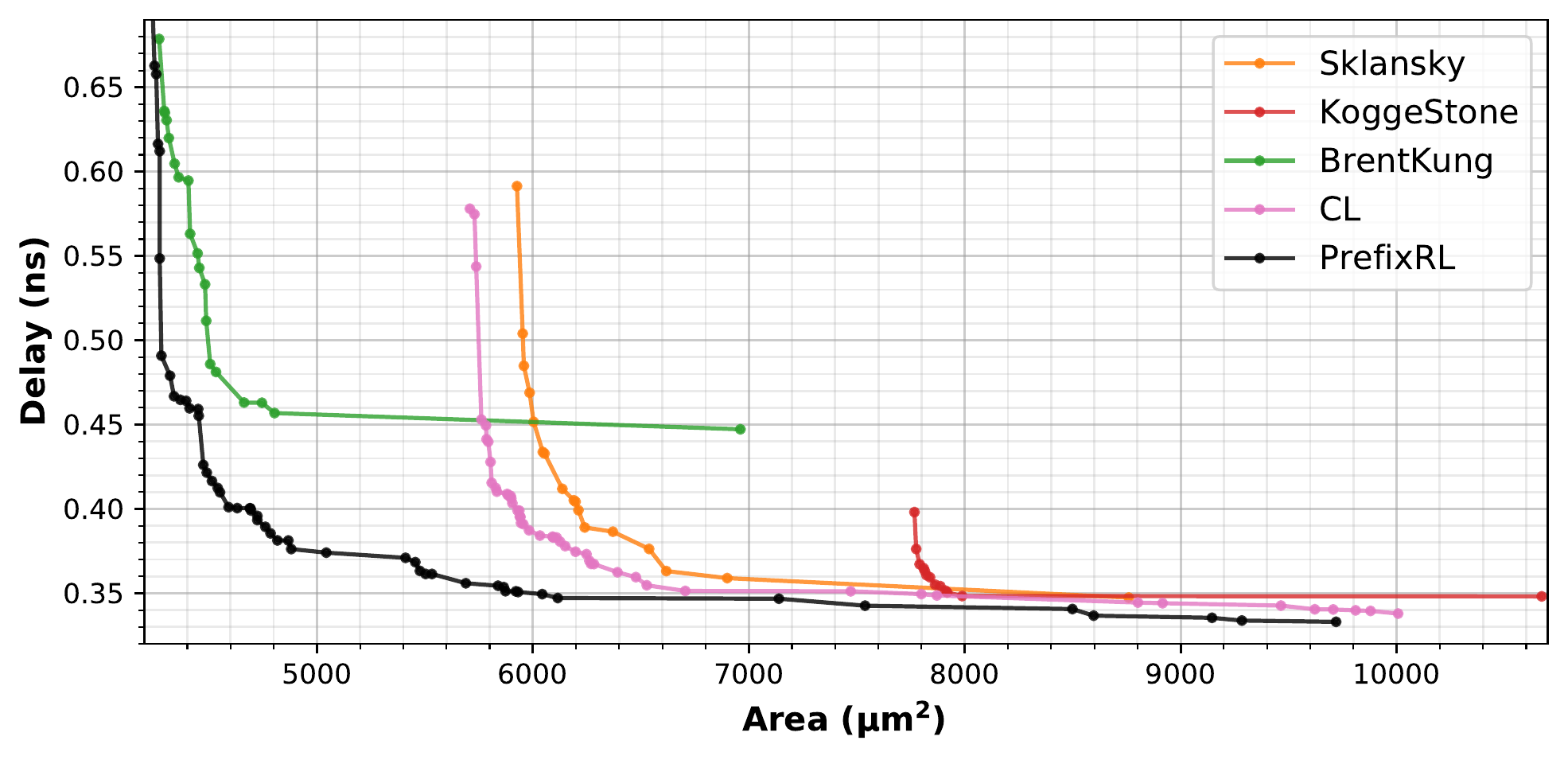}
        \caption{64b Adder Synthesis. OpenPhySyn, Nangate45.}
        \label{fig:openphysynresults64}
    \end{subfigure}%
        \caption{Area-delay Pareto curves for 32b and 64b adders synthesized with OpenPhySyn. PrefixRL adders Pareto-dominate all prior work.
        (Sklansky\cite{sklansky1960conditional}, KoggeStone\cite{kogge1973a}, BrentKung\cite{brent1982a}, SA\cite{moto2018prefix}, PS\cite{roy2014towards}, CL\cite{ma2019cross})
        }
        \label{fig:openphysynresults}
\end{figure}

%% file: figtex/fig_commercialresults.tex
\begin{figure}
    \setlength{\abovecaptionskip}{3pt}
    \setlength{\belowcaptionskip}{-10pt}
    \centering
    \begin{subfigure}[b]{\linewidth}
        \setlength{\abovecaptionskip}{0pt}
        \setlength{\belowcaptionskip}{0pt}
        \centering
        \includegraphics[width=\textwidth]{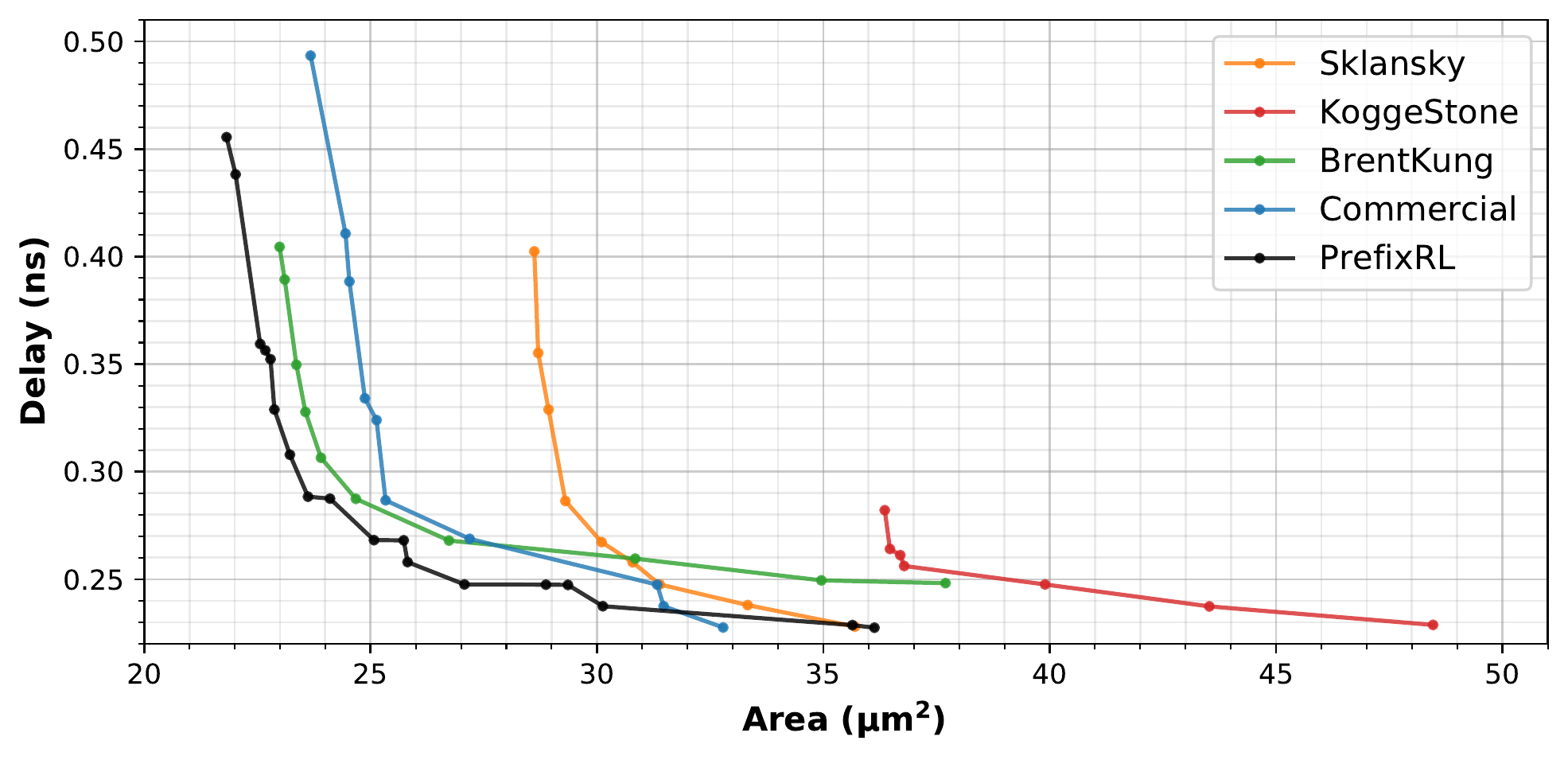}
        \caption{32b Adder Synthesis. Commercial tool, 8nm.}
        \label{fig:commercialresults32}
    \end{subfigure}%
    \hfill%
    \begin{subfigure}[b]{\linewidth}
        \setlength{\abovecaptionskip}{0pt}
        \setlength{\belowcaptionskip}{0pt}
        \centering
        \includegraphics[width=\textwidth]{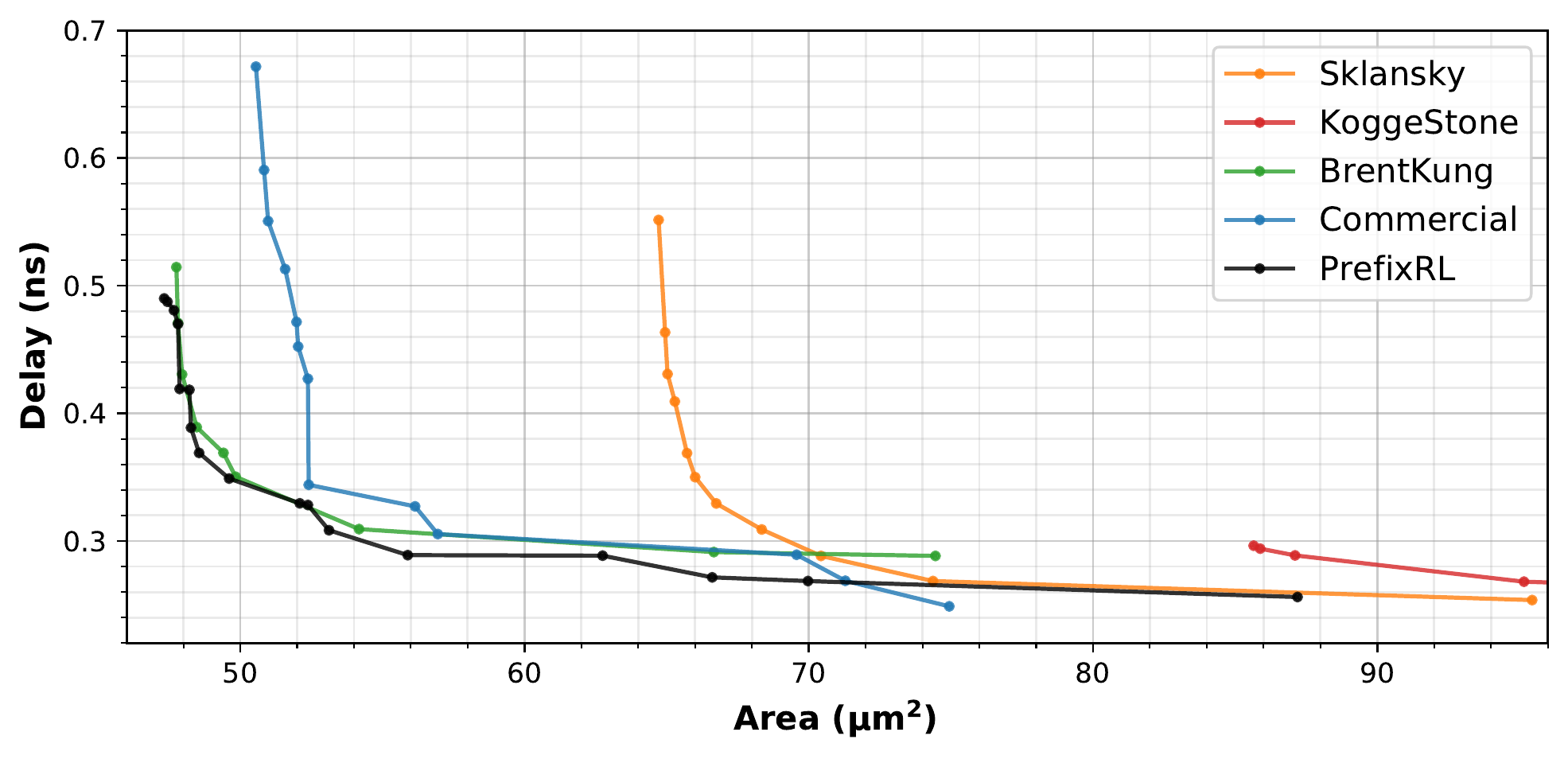}
        \caption{64b Adder Synthesis. Commercial tool, 8nm.}
        \label{fig:commercialresults64}
    \end{subfigure}%
        \caption{Area-delay Pareto curves for 32b and 64b adders synthesized using a Commercial Synthesis Tool. Despite having been trained only with OpenPhySyn, PrefixRL-generated adders still Pareto-dominate existing baselines at all but the lowest delay target.
        (Sklansky\cite{sklansky1960conditional}, KoggeStone\cite{kogge1973a}, BrentKung\cite{brent1982a})}
        \label{fig:commercialresults}
\end{figure}

%% file: figtex/fig_scaling.tex
\begin{table}[htbp]
  \centering
  \caption{Comparison of 16b, 32b and 64b PrefixRL adder design}
    \resizebox{\columnwidth}{!}{\begin{tabular}{cccc}
    \textbf{Statistic} & \multicolumn{1}{M{5.145em}}{\textbf{16b}} & \multicolumn{1}{M{5em}}{\textbf{32b}} & \multicolumn{1}{M{5.785em}}{\textbf{64b}} \\
    \toprule
    \textbf{$\vert \mathcal{A} \vert$} & 105 & 465 & 1953 \\
    \textbf{Synthesis time} & 11.39s & 16.85s & 35.56s \\ 
    \textbf{Train iteration time} & 0.45s & 1.61s & 3.15s \\
    \textbf{\# of residual blocks} & 16 & 32 & 32 \\
    \textbf{per-GPU batch size} & 96 & 96 & 6 \\
    \textbf{\# of data-parallel GPUs} & 1 & 1 & 14 \\
    \bottomrule
    \end{tabular}}%
\begin{tablenotes}[flushleft]
\item 
Synthesis times are for Sklansky adders evaluated at 4 timing constraints. The problem space grows quickly with the number of bits and impacts other details of training.  
\end{tablenotes}
  \label{fig-scaling}%
\end{table}%

%% file: figtex/fig_analyticalresults.tex
\begin{figure}[htbp]
    \setlength{\belowcaptionskip}{0pt}
    \centering
    \begin{subfigure}[b]{0.5\linewidth}
        \setlength{\abovecaptionskip}{0pt}
        \setlength{\belowcaptionskip}{0pt}
        \centering
        \includegraphics[width=\textwidth]{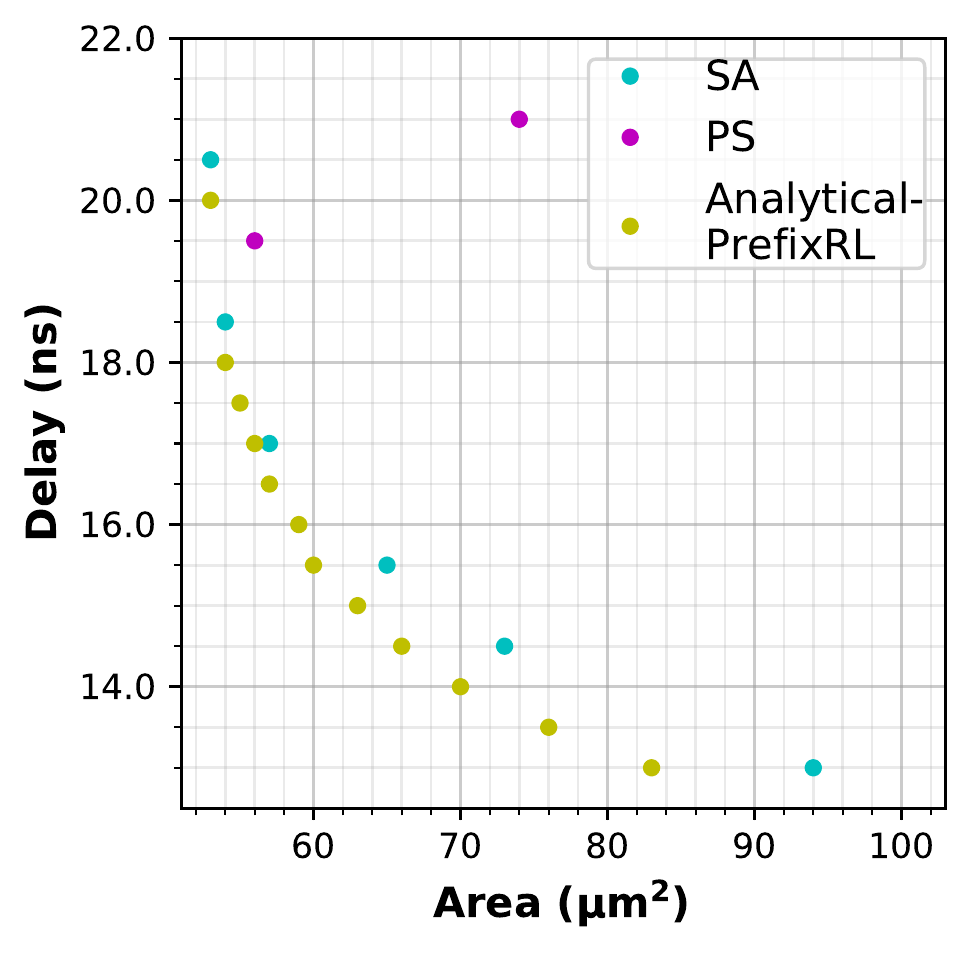}
         \caption{Analytical Metrics.}
        \label{fig:32anaresultsana}
    \end{subfigure}%
    \hfill%
    \begin{subfigure}[b]{0.5\linewidth}
        \setlength{\abovecaptionskip}{0pt}
        \setlength{\belowcaptionskip}{0pt}
        \centering
        \includegraphics[width=\textwidth]{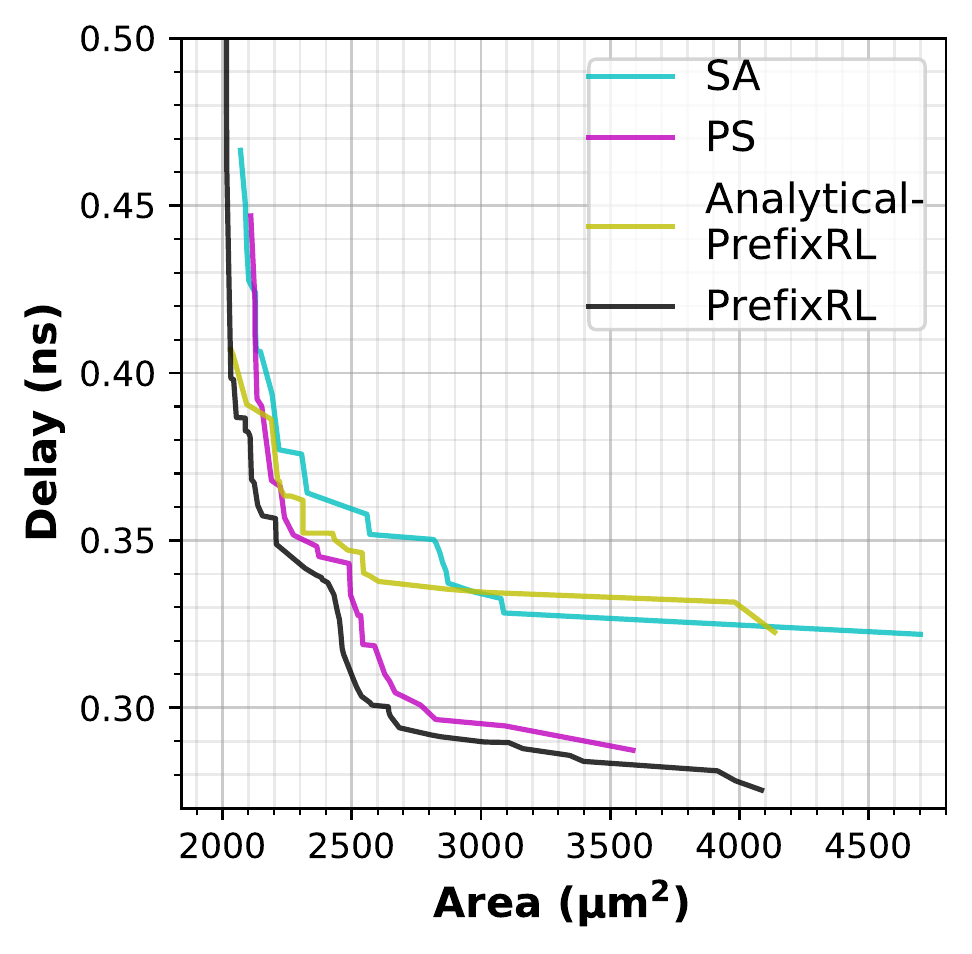}
        \caption{Synthesized Metrics.}
        \label{fig:32anaresultssynth}
    \end{subfigure}%
        \caption{32b Analytical-PrefixRL adders outperform SA\cite{moto2018prefix} and PS\cite{roy2014towards} adders when trained and compared with analytical metrics from \cite{moto2018prefix}. However, timing-driven synthesis optimizations in OpenPhySyn optimize the PS adders better than the Analytical-PrefixRL and SA adders in synthesized circuits. PrefixRL adders trained with OpenPhySyn synthesis in-the-loop have the best performing synthesized circuits.}
        \label{fig:32anaresults}
\end{figure}

%% file: 5_conclusion.tex
\section{Conclusion}
\label{sec:conclusions}
In this paper, we have presented PrefixRL: a new deep reinforcement learning based solution for optimizing prefix circuits for synthesized area and delay. PrefixRL does not use heuristics to evaluate solutions in a pruned design space, but rather uses a deep neural network model that learns strategies purely through exploration of the unrestricted design space and feedback from synthesis tools.
We apply PrefixRL to the task of designing area-delay optimized 32b and 64b prefix adders and demonstrate that it finds a frontier of designs across various area-delay trade-offs that significantly outperform previous methods. We observe that agents trained with open-source synthesis tools and cell library can design adder circuits that achieve lower area and delay than commercial tool adders in an industrial cell library. This result further suggests that using target cell libraries or commercial synthesis tools during training is a promising direction for further improvement.
PrefixRL demonstrates the potential for deep reinforcement learning as an effective optimization algorithm for prefix circuits. In the future, we hope to extend the framework to other datapath circuits, consider nonuniform timing constraints and power objectives.

\begin{figure}[htbp]
\setlength{\abovecaptionskip}{3pt}
\setlength{\belowcaptionskip}{0pt}
\centerline{\includegraphics[width=0.8\linewidth]{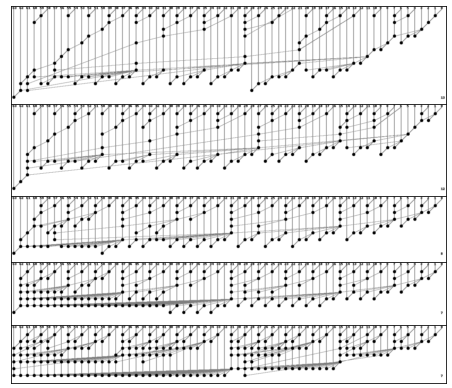}}
\caption{64b PrefixRL Solutions}
\label{fig:64solutions}
\end{figure}